\documentclass[Journal, NoLineNumbers,letterpaper]{ascelike-new}
\usepackage[utf8]{inputenc}
\usepackage[T1]{fontenc}
\usepackage{lmodern}
\usepackage{graphicx}
\usepackage[figurename=Fig.,labelfont=bf,labelsep=period]{caption}
\usepackage{subcaption}
\usepackage{amsmath}
\usepackage{newtxtext,newtxmath}
\usepackage[colorlinks=true,citecolor=red,linkcolor=black]{hyperref}
\NameTag{Ghasemi, \today}
\begin{document}

\title{Improving the Accuracy Of MEPDG Climate Modeling Using Radial Basis Function}

\author[1]{Amirehsan Ghasemi}
\author[2]{Kelvin J Msechu}
\author[3]{Arash Ghasemi}
\author[4]{Mbakisya A. Onyango}
\author[5]{Joseph Owino}
\author[6]{Ignatius Fomunung}

\affil[1]{University of Tennessee at Chattanooga, Tennessee, USA. amirehsan-ghasemi@mocs.utc.edu}
\affil[2]{University of Tennessee at Chattanooga, Tennessee, USA}
\affil[3]{University of Tennessee at Chattanooga, Tennessee, USA}
\affil[4]{University of Tennessee at Chattanooga, Tennessee, USA}
\affil[5]{University of Tennessee at Chattanooga, Tennessee, USA}
\affil[6]{University of Tennessee at Chattanooga, Tennessee, USA}

\maketitle

\begin{abstract}
In this paper, the accuracy of two mesh-free approximation approaches, the Gravity model and Radial Basis Function, are compared. The two schemes' convergence behaviors prove that RBF is faster and more accurate than the Gravity model. As a case study, the interpolation of temperature at different locations in Tennessee, USA, are compared. Delaunay mesh generation is used to create random points inside and on the border, which data can be incorporated in these locations. 49 MERRA weather stations as used as data sources to provide the temperature at a specific day and hour. The contours of interpolated temperatures provided in the result section assert RBF is a more accurate method than the Gravity model by showing a smoother and broader range of interpolated data.
\end{abstract}

\section{Introduction}

In current data driven world, maintaining the quality of data and its derivatives is of most importance. Complete data sets are necessary to properly understand, interpret and use data for various applications; however, gaps do exist in data. One of the methods employed to fill data gaps is data interpolation. The purpose of this study was to assesses the two common methods used in data interpolation; gravity model and radial basis function model. Both models are still used in different applications, but their output accuracies are yet to be compared. When comparing both  interpolation models’ outputs to the actual analytical outputs, the radial basis function model achieved  faster, and  higher accuracy as compared to the gravity model. To illustrate the effects of these interpolated output differences, temperature contour maps were produced for the state of Tennessee considering; 49 MERRA weather stations as data sources and, 3881 data interpolation points. The interpolated temperature contours were much smoother, and well defined with the RBF model and it’s output ranges were wider than the gravity model.

When dealing with  , being a small data set or big data, quality is the most important factor to consider. In the growing demand of tools and methods to deal with the vast daily or periodic data, it is important to consider tools and methods that will provide desired data quality. For cases where data gap(s) are experienced, different methods can be employed to fill these gaps. Filling of data gaps is necessary for multiple tasks such as; accurate data interpretations, model calibrations and realistic presentation of information. One common method  used in filling empty gaps while considering existing data quantities is data interpolation. Many interpolation methods have been developed throughout the years with different applications in various fields of study. Common to these widely used interpolation methods include the gravity model and the radial basis function (RBF) model. Being the common methods for data interpolation, it is vital to determine the accuracy offered by each method.

\section{Gravity Model}

Inspired by the Newton’s universal law of gravity that relates the gravitational force between two bodies to the product of their masses and being inversely proportional to the square of the distance between them, the gravity model also applies the same principle. The gravity model has different applications throughout multiple disciplines including the transportation industry, civil engineering, data and computer science fields.
Some application of the gravity model includes data classification \cite{6403569}. Gravity model algorithms have been used for multi-label lazy learning, by treating instances of data as  particles of the data. The use of gravity model in multi-label lazy learning reported better performance than the state-of-the-art multi-label lazy methods \cite{REYES2016159}. The gravity models in combination with cognitive laws have been used for small noisy data classification \cite{WEN2013245}. Gravity models have been employed to improve the local weighted learning regression method for multi-target data \cite{Reyes2018}.
For data interpolation purposed, the gravity model is being used in the formation of Visual Weather Station (VWS)  for the mechanistic-empirical pavement design procedures \cite{Schwartz1}.
Gravity Model Interpolation theory
The gravity interpolation model also known as 1/R method considers distances in establishing weights for its final interpolated output. During the interpolation, closest points to the interpolation point of interest, contributes more weight to the final output. The gravity model uses the equation as seen below. Where Dint is the gravity interpolation output, D is the data that is obtained from the different neighboring points at  distance, (Eq.~\ref{eq:gravity_model}) \cite{Schwartz1}.

\begin{equation} \label{eq:gravity_model}
D_{int} = \frac{\sum_{i=1}^{n} \frac{D_i}{d_i^2}}{\sum_{i=1}^{n} \frac{1}{d_i^2}}
\end{equation}

\section{Radial Basis Function}

The general linear governing equation for RBF can be expressed as

\begin{equation} \label{eq:rbf}
h(x) = \sum_{i=1}^{N} \omega_i \phi(||x_i-x_a||)
\end{equation} where $\omega$ and $\phi$ represent the weights and basis function of the network, respectively. Basis functions $(\phi)$ are the function of distance for $N$ number of points form a specific fixed point $x_a$. For finding the distance, several different functions can be considered. In this paper Euclidean distance function is used, which can be shown as $||x_i-x_a|| = \sqrt{(x_i-x_a)^2+(y_i-y_a)^2}$. The distance can be denoted as $r$ so the (Eq.~\ref{eq:rbf}) may be written as

\begin{equation} \label{eq:rbf_r}
h(x) = \sum_{i=1}^{N} \omega_i \phi(r)
\end{equation}

There are some forms of $\phi(r)$ to choose from. The following are some of the famous basis functions which are commonly used

\begin{equation} \label{eq:Gaussian}
\textrm{Gaussian:}             \qquad          \phi(r) = e^{-\varepsilon r^2}
\end{equation}

\begin{equation} \label{eq:Multiquadric}
\textrm{Multiquadric:}         \qquad          \phi(r) = \sqrt{1+(\varepsilon r)^2}
\end{equation}

\begin{equation} \label{eq:Inverse_multiquadratic}
\textrm{Inverse Multiquadric:} \qquad          \phi(r) = \frac{1}{\sqrt{1+(\varepsilon r)^2}}
\end{equation}

\begin{equation}  \label{eq:Thin_plate_spline}
\textrm{Thin Plate Spline:}    \qquad          \phi(r) = r^2\ln(r)
\end{equation} where $\varepsilon$ is a positive scale parameter known as the shape parameter and helps put a smooth surface to the data.

\subsection{RBF: METHODOLOGY}

(Fig.~\ref{fig:mth_points}) shows $N$ number of points. These points have a value which can be considered as $h$. If one of these points, for example, point 1, is chosen as a fixed point and has a value as $h_1$, then  (Eq.~\ref{eq:rbf}) for point 1 can be expressed as

\begin{equation}  \label{eq:point1}
 h_1 = \omega_1 \phi(||x_1-x_1||) + \omega_2 \phi(||x_2-x_1||) + \cdots + \omega_N \phi(||x_N-x_1||) 
\end{equation} by considering each point from all $N$ number of points as a fixed point and using (Eq.~\ref{eq:rbf}), the following system of equation is gained

\begin{equation} \label{eq:sys}
\begin{aligned}
& h_1 = \omega_1 \phi(||x_1-x_1||) + \omega_2 \phi(||x_2-x_1||) + \cdots + \omega_N \phi(||x_N-x_1||)\\    
& h_2 = \omega_1 \phi(||x_1-x_2||) + \omega_2 \phi(||x_2-x_2||) + \cdots + \omega_N \phi(||x_N-x_2||)\\  	
& h_3 = \omega_1 \phi(||x_1-x_3||) + \omega_2 \phi(||x_2-x_3||) + \cdots + \omega_N \phi(||x_N-x_3||)\\	
& h_N = \omega_1 \phi(||x_1-x_N||) + \omega_2 \phi(||x_2-x_N||) + \cdots + \omega_N \phi(||x_N-x_N||)\\		
\end{aligned}
\end{equation} expressing the (Eq.~\ref{eq:sys}) as a matrix notation, all the shape functions $\phi(r), $ is considered in one matrix, which is known as interpolation matrix and is non-singular

\begin{equation} \label{eq:interpolation_mat}
\left[
\begin{array}{c c c c}
	\phi(||x_1-x_1||)     &     \phi(||x_2-x_1||)      &     \cdots      &      \phi(||x_N-x_1||)   \\
  \phi(||x_1-x_2||)     &     \phi(||x_2-x_2||)      &     \cdots      &      \phi(||x_N-x_2||)   \\
  \vdots                &     \vdots                 &     \ddots      &      \vdots              \\
  \phi(||x_1-x_N||)     &     \phi(||x_2-x_N||)      &     \cdots      &      \phi(||x_N-x_N||)   \\
\end{array}
\right]
\end{equation} with having the matrices of interpolation and matrix of $h$, which is the value for each point, the values for $\omega_1$ to $\omega_N$ are found by solving the following linear system

\begin{equation} \label{eq:linear_sys}
\left[	
\begin{array}{c }
\omega_1    \\
\omega_2    \\
\vdots      \\
\omega_N 
\end{array}
\right]
=
\left[	
\begin{array}{c c c c}
\phi(||x_1-x_1||)     &     \phi(||x_2-x_1||)      &     \cdots      &      \phi(||x_N-x_1||)   \\
\phi(||x_1-x_2||)     &     \phi(||x_2-x_2||)      &     \cdots      &      \phi(||x_N-x_2||)   \\
\vdots                &     \vdots                 &     \ddots      &      \vdots              \\
\phi(||x_1-x_N||)     &     \phi(||x_2-x_N||)      &     \cdots      &      \phi(||x_N-x_N||)   
\end{array}
\right]^{-1}
\left[	
\begin{array}{c}
h_1    \\
h_2    \\
\vdots      \\
h_N 
\end{array}
\right]
\end{equation}

Once the weight ($\omega$) for all points are available, the approximation for any desirable fixed point is possible. For example, considering point $b$ as a fixed point, the approximed $h$ value for this point $(h_b)$ can be provided by using (Eq.~\ref{eq:rbf}) as follow

\begin{equation}  \label{eq:point1}
 h_b = \omega_1 \phi(||x_1-x_b||) + \omega_2 \phi(||x_2-x_b||) + \cdots + \omega_N \phi(||x_N-x_b||) 
\end{equation}

\section{Convergance Behaviour}

The most important aspect while dealing with data is to maintain and or obtain higher accuracy data. Through this section, the interpolation accuracy of the gravity model and the radial basis function model were studied by considering two groups of  generated numeric data. The first group included  a grid of X and Y values ranging from [0,1] with an equal interval spacing of 0.125 in both directions. Z value at each grid point were generated by an analytical cosine function (Eq.~\ref{eq:cosine}) with respect to the X-Y values of the points. The objective of the generated X - Y grid points were to serve as the  points for interpolation, where the outputs of the interpolation (Z-calculated) were to be compared to the analytical Z values  of the respective points.

\begin{equation}  \label{eq:cosine}
f = 2 + 0.2 \cos(2 \pi x)  \cos(2  \pi y)
\end{equation}

Like the first group, group 2 values of the x and y points are randomly generated (non-gridded) from the range of [0,1] in both x and y directions. The z value for each random point were generated from the analytical cosine function as done for group 1. Group 2’s x, y and z points were used as the interpolation data in both the models to obtain Z-calculated at the respective X and Y grid points.

The procedures to generate the Z-calculated values as previously described, was done starting from 300 to 15,000 random points at an interval of 300. For each number of  random points used the Root Mean Square Error (RMS) was computed comparing the analytical Z values with Z-calculated values from both models’ interpolations. The values obtain from the error analyses are plotted with the x-axis having the number of random points used (Points) versus the RMS error, calculated with respect to the number of random points used.

\begin{figure}
	\centering
	\includegraphics[trim=100 230 100 200,clip, width=\textwidth]{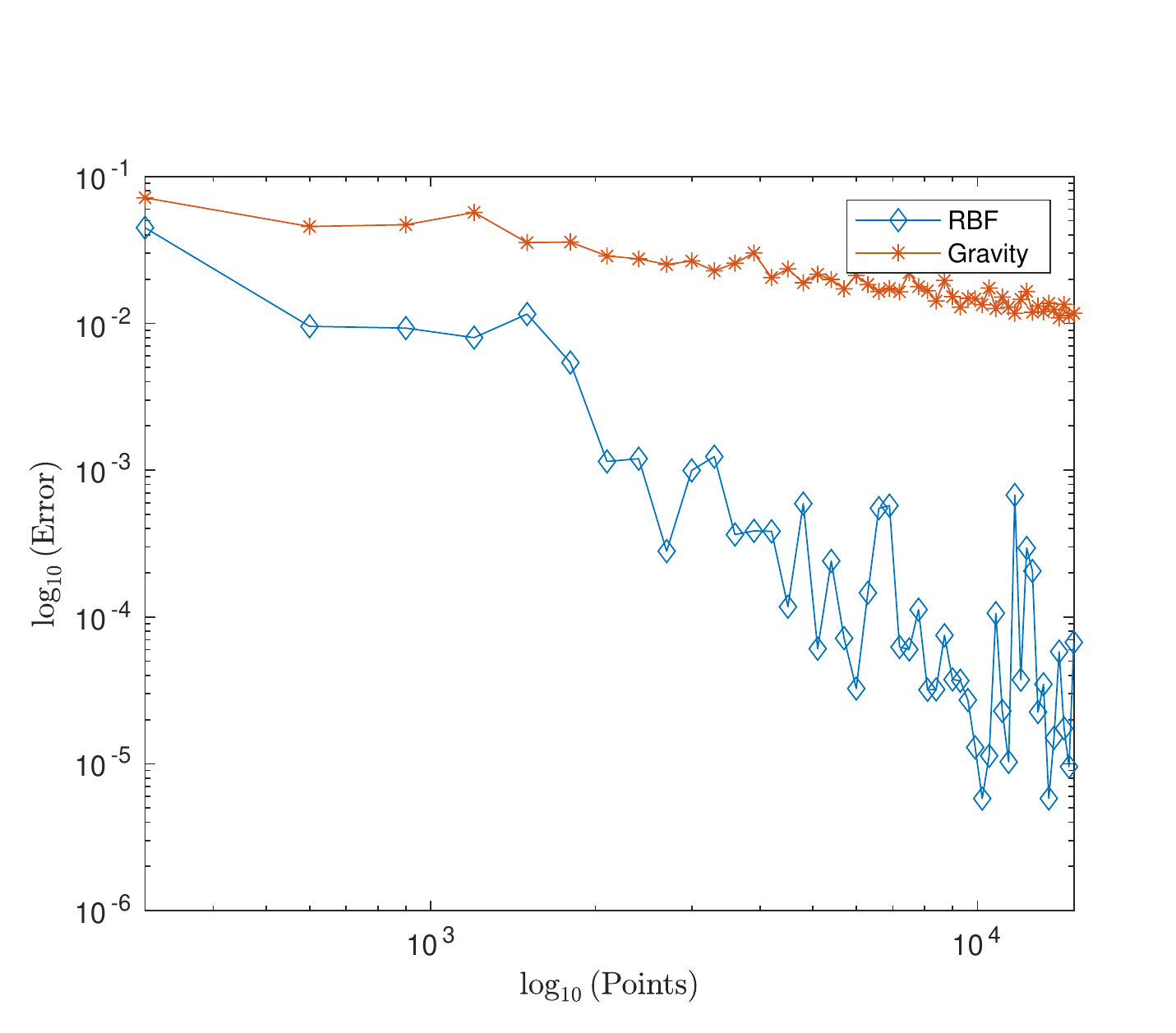}  
	\caption{Convergence Behavior Of The 1$^{\textrm{st}}$ Run}
	\label{fig:error1}
\end{figure}

\begin{figure}
	\centering
	\includegraphics[trim=100 230 100 200,clip, width=\textwidth]{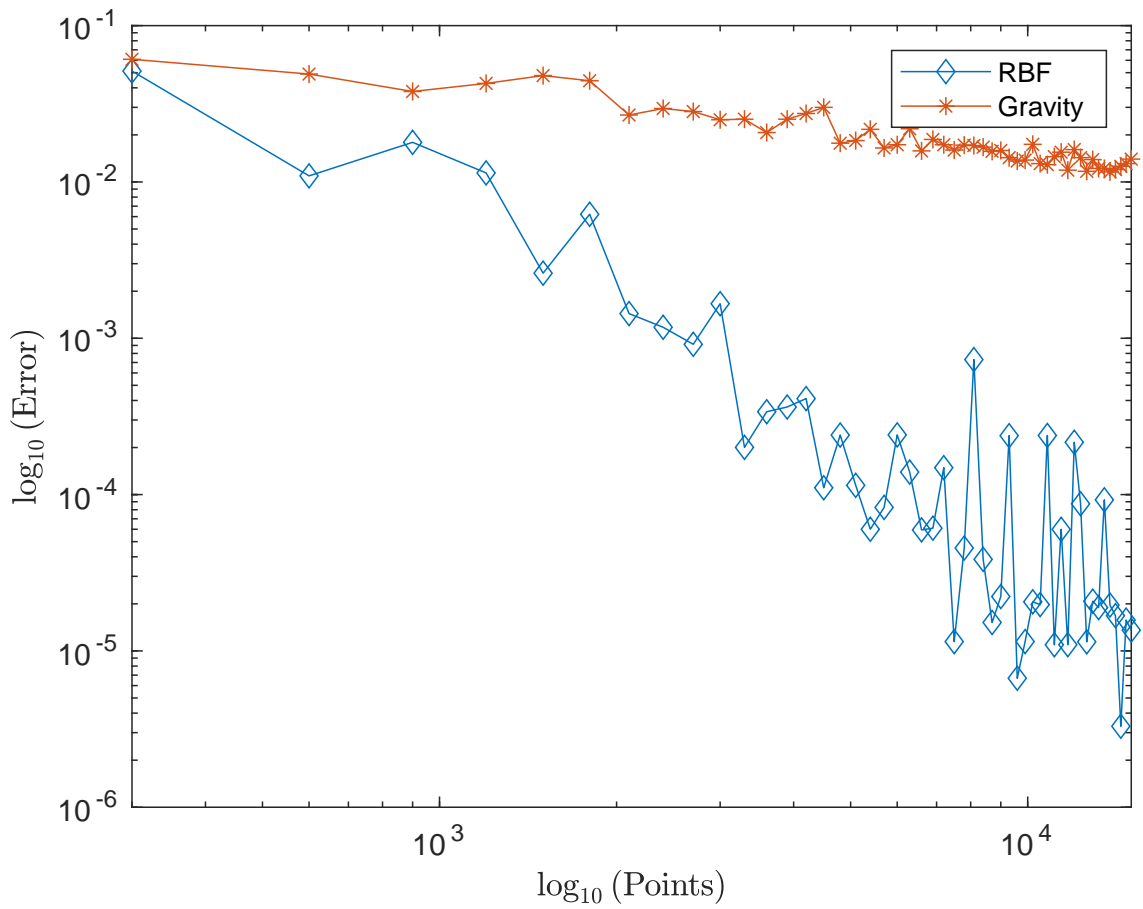}  
	\caption{Convergence Behavior Of The 2$^{\textrm{nd}}$ Run}
	\label{fig:error2} 
\end{figure}

\begin{figure}
	\centering
	\includegraphics[trim=100 230 100 200,clip, width=\textwidth]{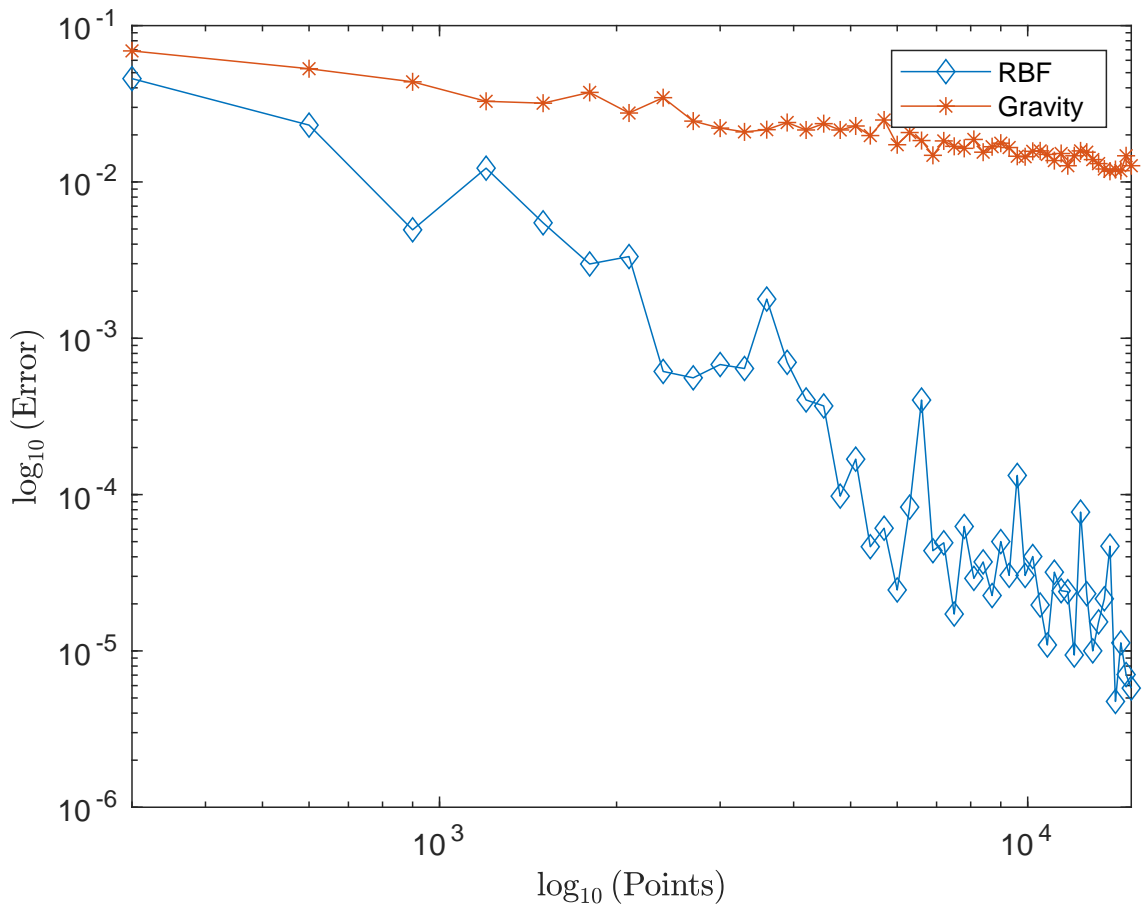}  
	\caption{Convergence Behavior Of The 3$^{\textrm{rd}}$ Run}
	\label{fig:error3}
\end{figure}

\begin{figure}
	\centering
	\includegraphics[trim=100 230 100 200,clip, width=\textwidth]{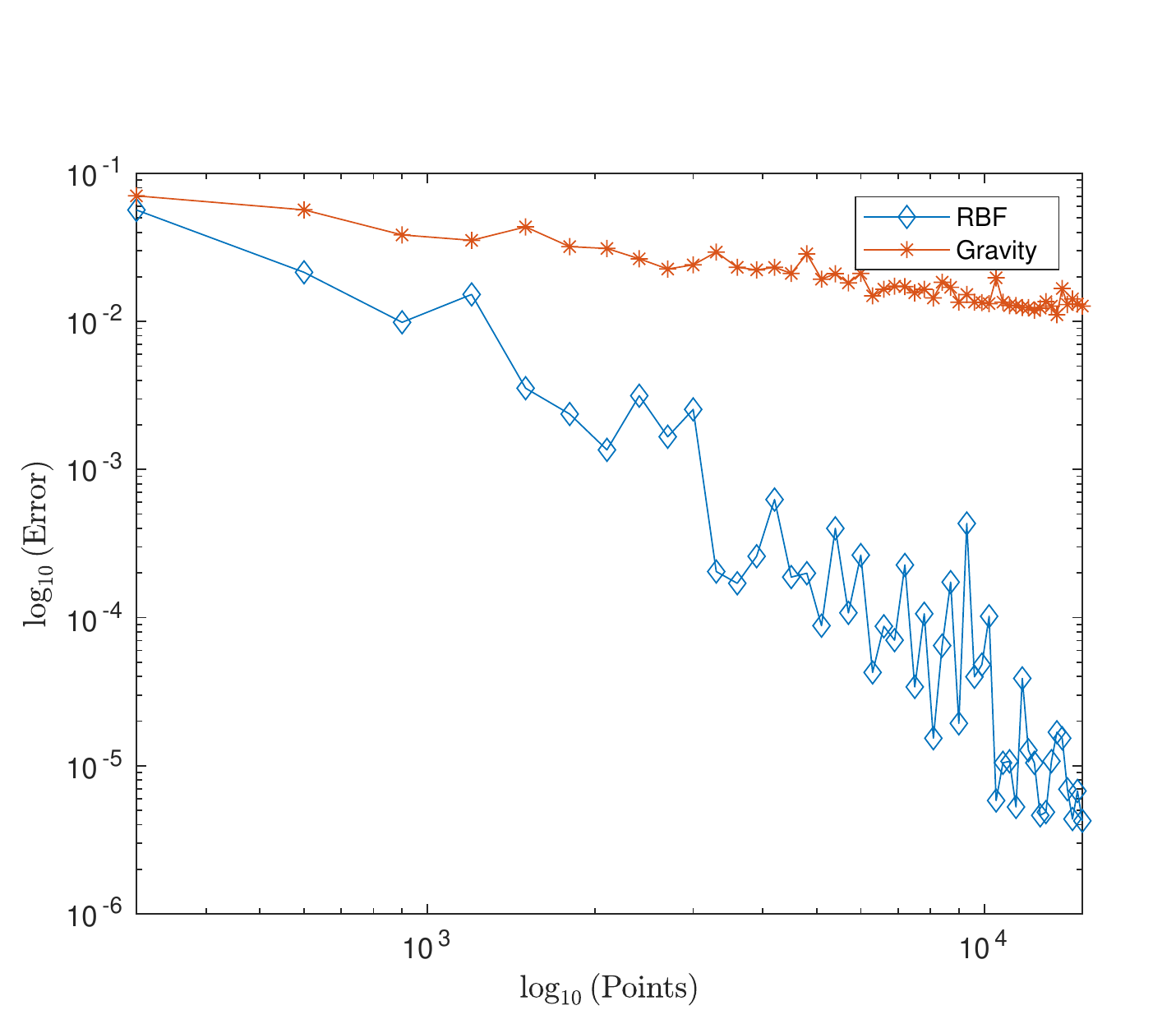}  
	\caption{Convergence Behavior Of The 4$^{\textrm{th}}$ Run}
	\label{fig:error4}
\end{figure}

\begin{figure}
	\centering
	\includegraphics[trim=100 230 100 200,clip, width=\textwidth]{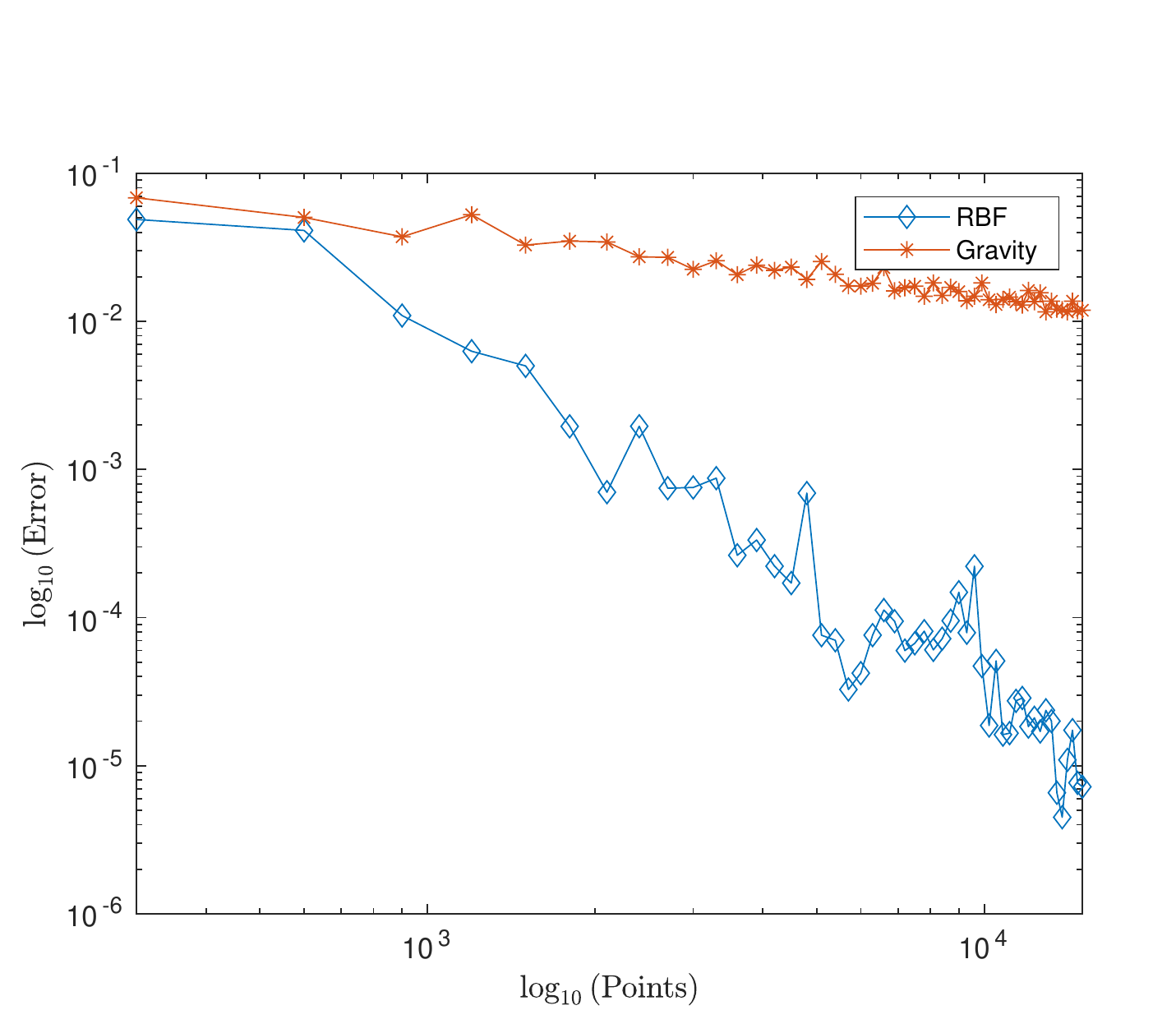}  
	\caption{Convergence Behavior Of The 5$^{\textrm{th}}$ Run}
	\label{fig:error5}
\end{figure}

(Fig.~\ref{fig:error1}) to (Fig.~\ref{fig:error5})show the  trend of the RBF interpolation output when the RMS error values are plotted against the number of random points used in the interpolation. As From Figures show, the trend of the RBF errors is observed to have a steep negative slope as the number or interpolation points increase so the RBF interpolation considered the more accurate. It is evident that the  RBF’s plot achieved lower RMS error values faster than the gravity model with respect to the number of random points. The gravity model plot, RMS error decreases slowly with the increase in points. From the observation the RBF interpolation proves to be more robust than the gravity model, requiring less number of data points to achieve similar or higher accuracy.

\section{Case Study: Tennessee}

As a demonstration of accuracy differences, presented are  visual outputs for both the two interpolation techniques while considering Modern Era Rectospective-anaysis for Research and Application (MERRA) weather stations. The MERRA weather stations have a broad use from climate studies to application in civil engineering pavement design practices. Spatial information and climatic data are obtained from the Long-Term Pavement Performance (LTPP) InfoPave website.

\begin{figure}
	\centering
	\includegraphics[trim=100 230 100 200, width=\textwidth]{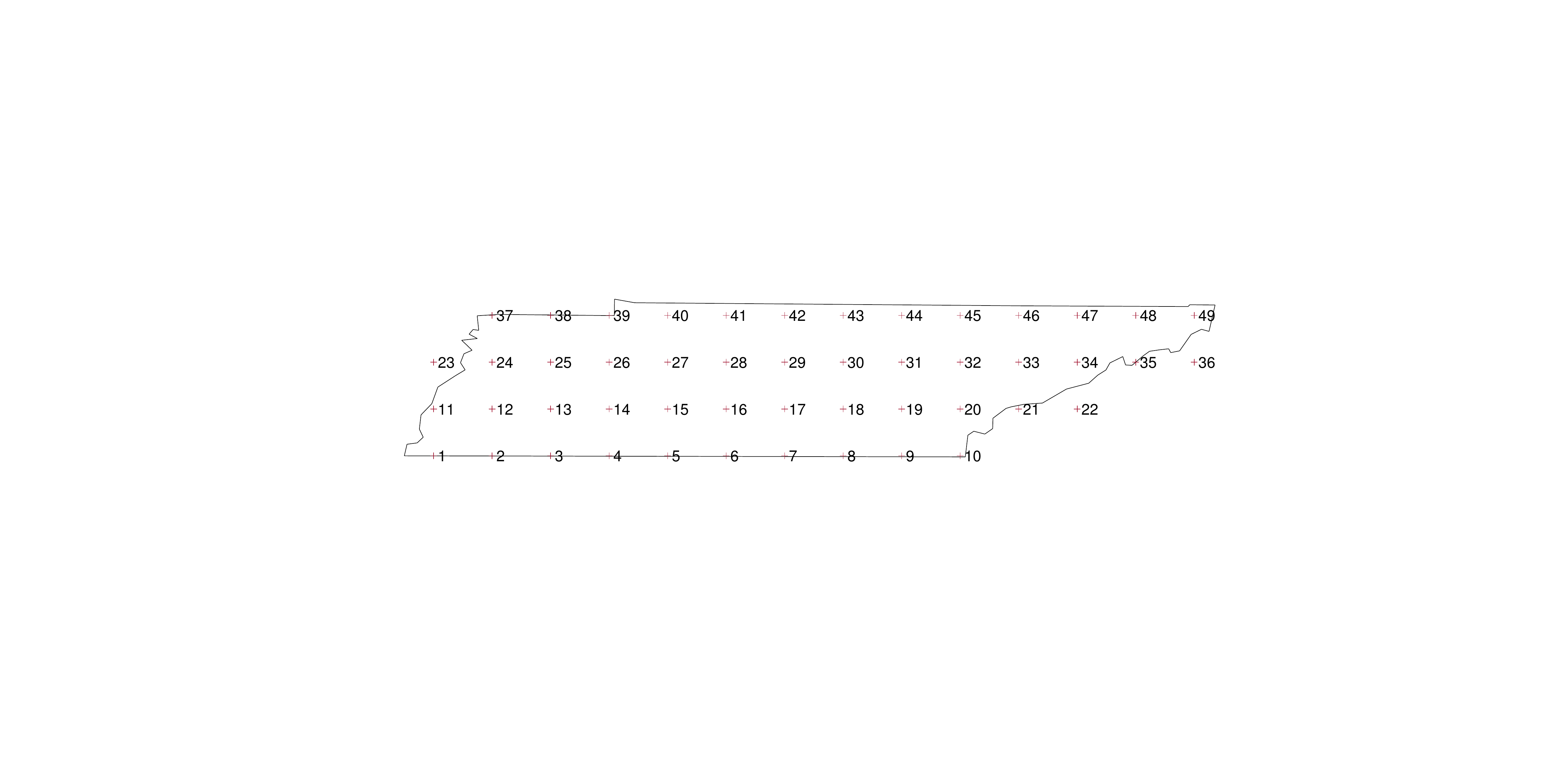}  
	\caption{Tennessee State Border and 49 MERRA Weather Stations}
	\label{fig:tn_border}
\end{figure}

Using the state of Tennessee as a case study, 49 MERRA climatic weather stations at a spatial resolution of 0.625o longitude by 0.5o latitude (Figure~\ref{fig:tn_border}) were used as the interpolation data sources. The MERRA data files contain hourly climatic data (hcd); temperature (°F), wind speed (mph.), percent sunshine, precipitation (inches.) and percent humidity. For this visual comparison of the interpolation methods, only temperature values were considered. Since MERRA data are recorded in hourly basis, only the first, 50,000, 100,000, 150,000 , 200,000 and 250,000 values were considered independently from each of the 49 MERRA climatic station for interpolation.

\begin{figure}
	\centering
	\includegraphics[trim=100 230 100 200, width=\textwidth]{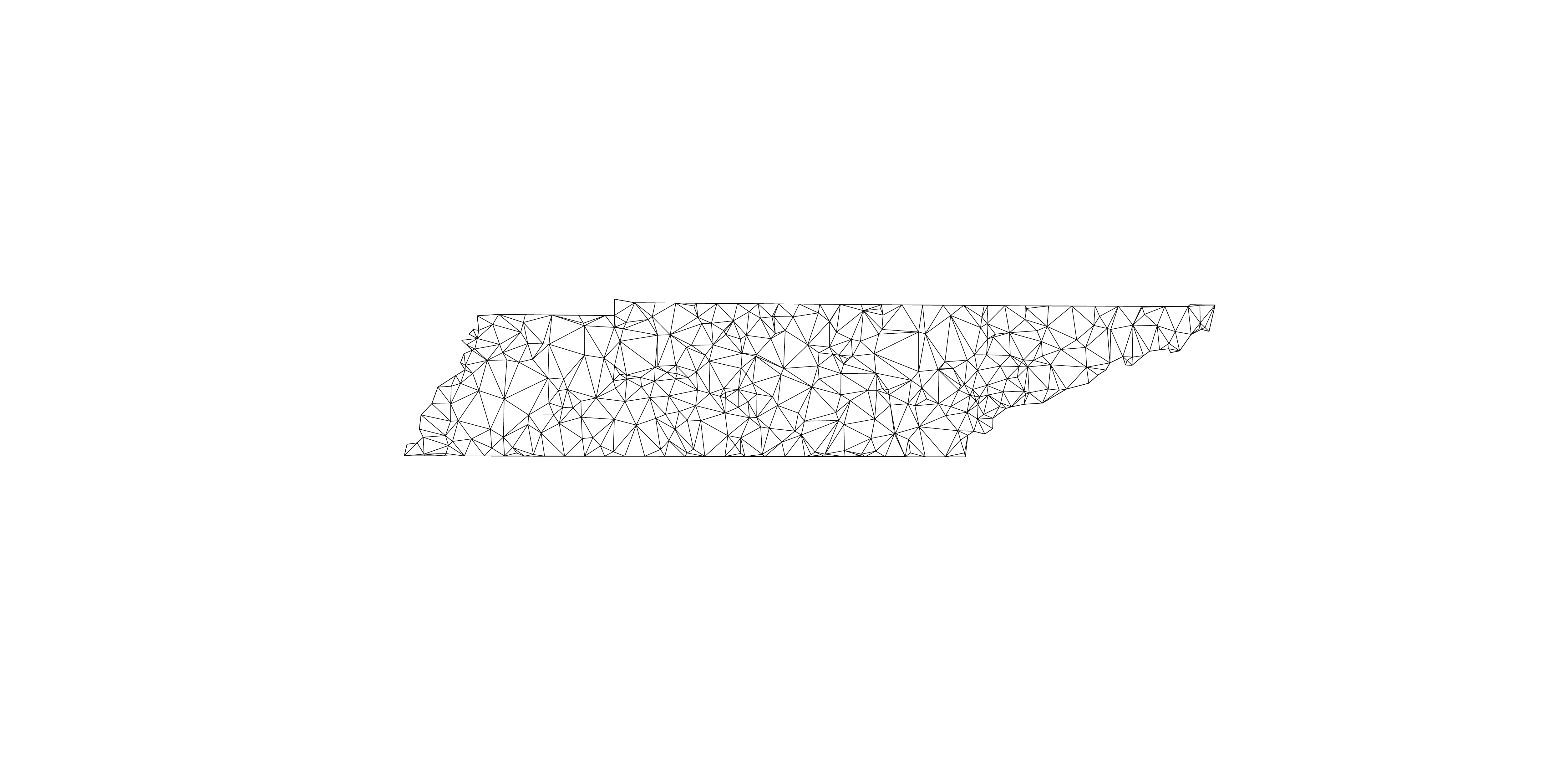}  
	\caption{Mesh Generation}
	\label{fig:mesh}
\end{figure}

\begin{figure}
	\centering
	\includegraphics[trim=0 0 0 0,clip, width=\textwidth]{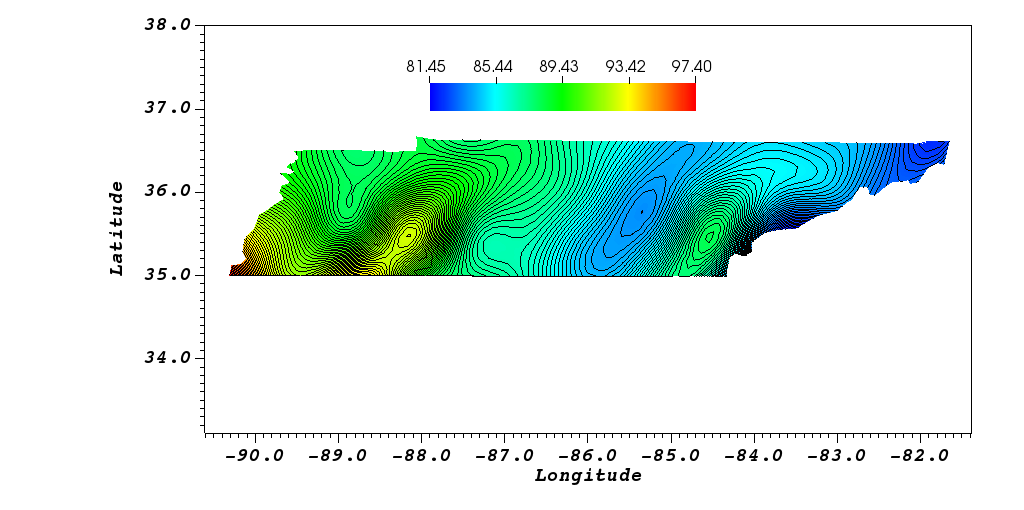}  
	\caption{RBF: Hour 6}
	\label{fig:rbf6}
\end{figure}

The state of Tennessee was meshed to obtain set of data with 3881 points (Figure~\ref{fig:mesh}), these points served as the points for interpolation. Gravity and RBF model interpolation were executed for each set of points (3881) using the 49 MERRA climatic stations as data sources. Temperature outputs from the interpolation at all point in each set of points were plotted as contours in Tennessee. As observed on (Figure~\ref{fig:gavity1}) to (Figure~\ref{fig:rbf6}), the visualized contours of gravity model interpolated temperature output have a less smooth and well-defined contours as the RBFs. It can also be observed that the gravity model produced higher low temperature values and lower high temperature values than RBF. This means that the gravity model interpolation outputs have a smaller range than the respective RBF model.

\begin{figure}
	\centering
	\includegraphics[trim=0 0 0 0,clip, width=\textwidth]{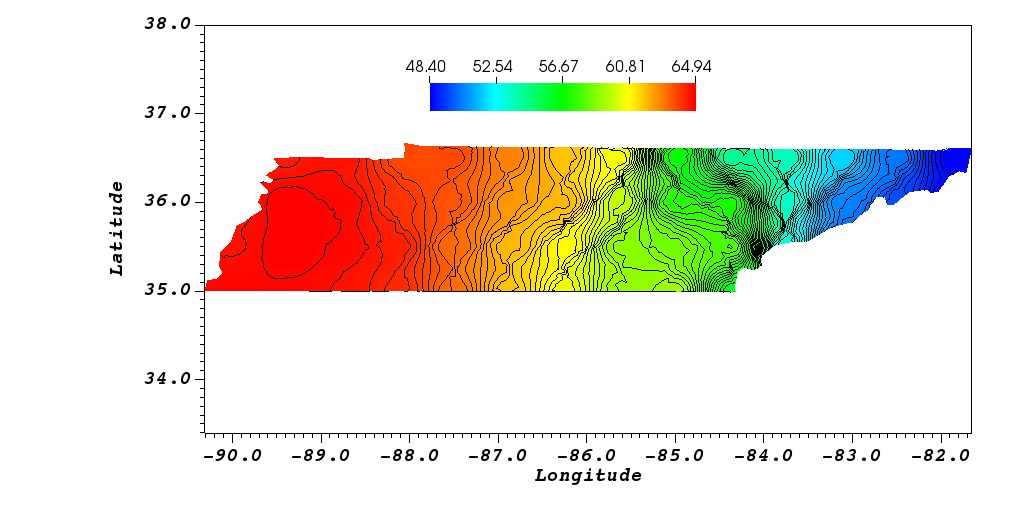}  
	\caption{Gravity Model: Hour 1}
	\label{fig:gravity1}
\end{figure}

\begin{figure}
	\centering
	\includegraphics[trim=0 0 0 0,clip, width=\textwidth]{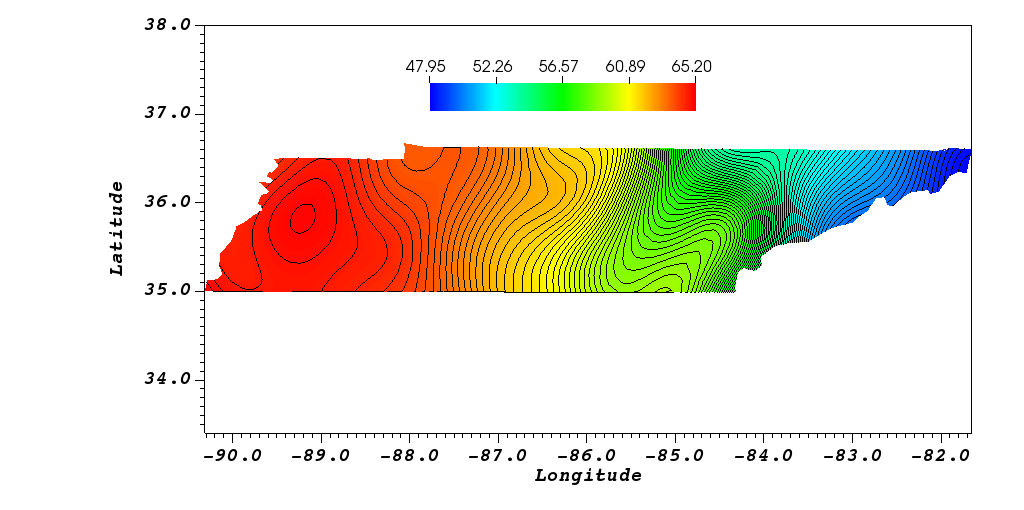}  
	\caption{RBF: Hour 1}
	\label{fig:rbf1}
\end{figure}

\begin{figure}
	\centering
	\includegraphics[trim=0 0 0 0,clip, width=\textwidth]{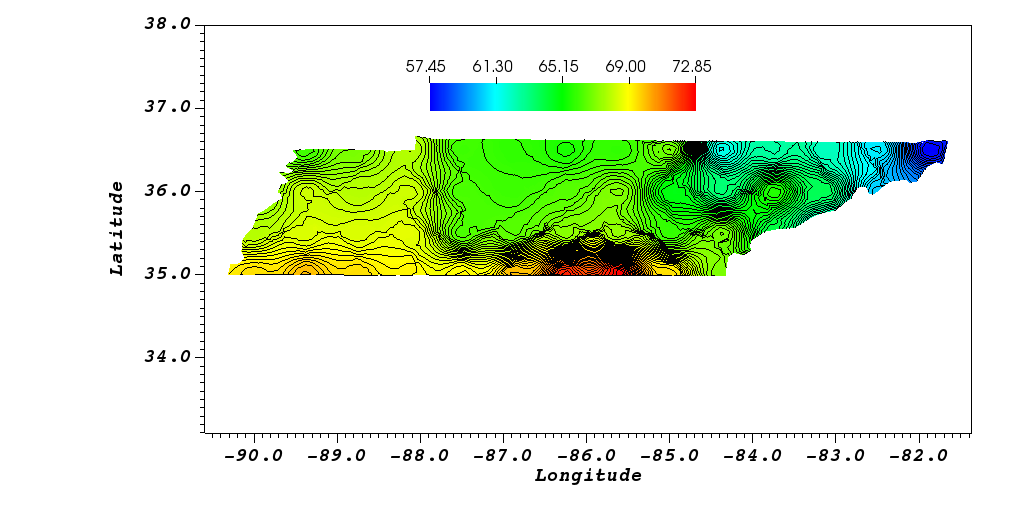}  
	\caption{Gravity Model: Hour 2}
	\label{fig:gravity2}
\end{figure}

\begin{figure}
	\centering
	\includegraphics[trim=0 0 0 0,clip, width=\textwidth]{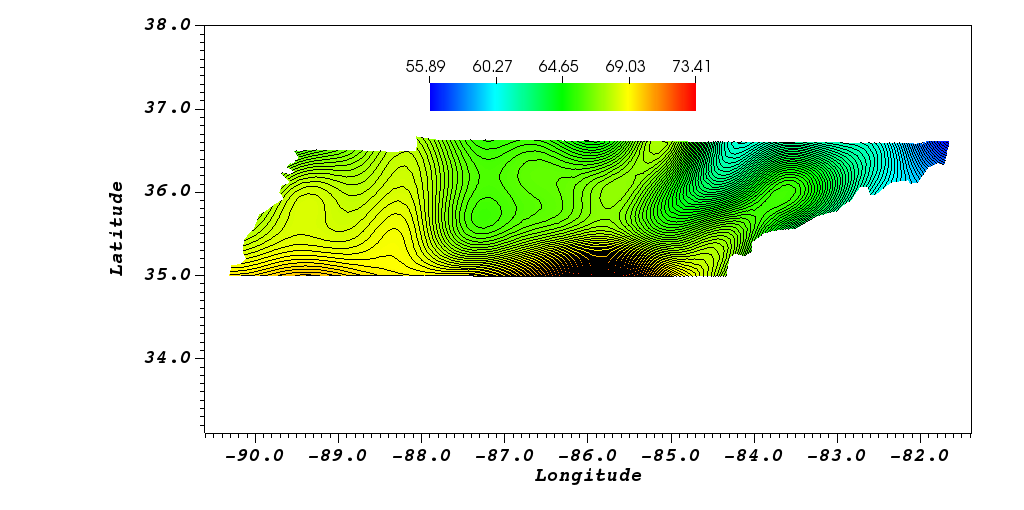}  
	\caption{RBF: Hour 2}
	\label{fig:rbf2}
\end{figure}

\begin{figure}
	\centering
	\includegraphics[trim=0 0 0 0,clip, width=\textwidth]{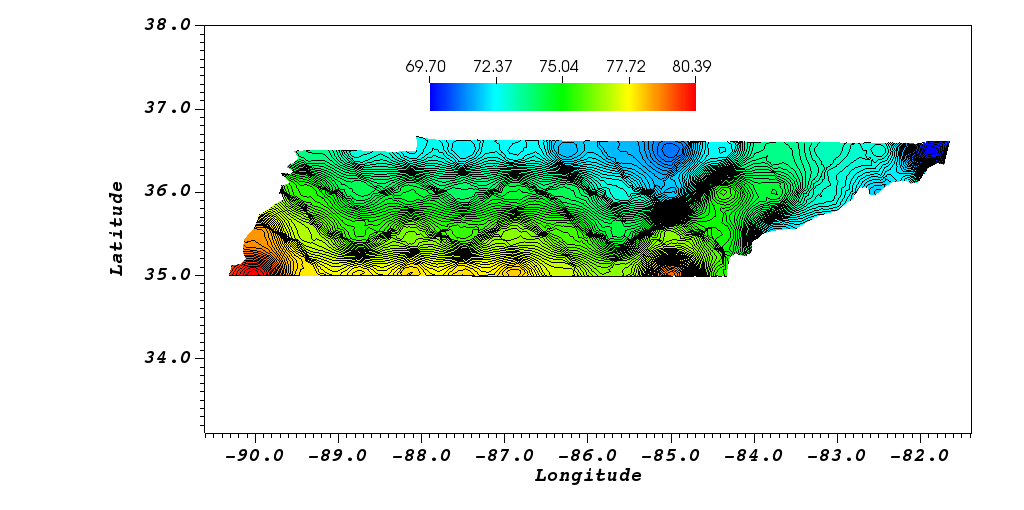}  
	\caption{Gravity Model: Hour 3}
	\label{fig:gravity3}
\end{figure}

\begin{figure}
	\centering
	\includegraphics[trim=0 0 0 0,clip, width=\textwidth]{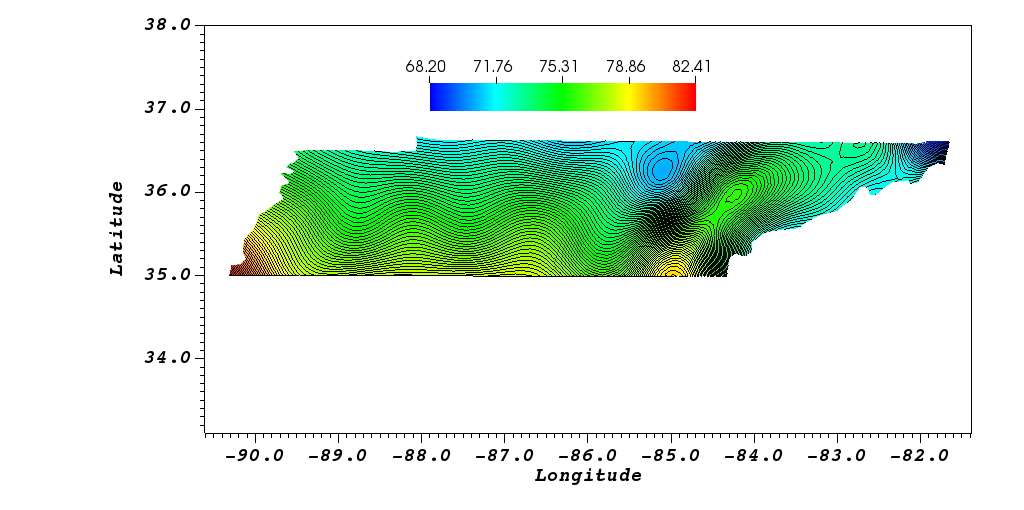}  
	\caption{RBF: Hour 3}
	\label{fig:rbf3}
\end{figure}

\begin{figure}
	\centering
	\includegraphics[trim=0 0 0 0,clip, width=\textwidth]{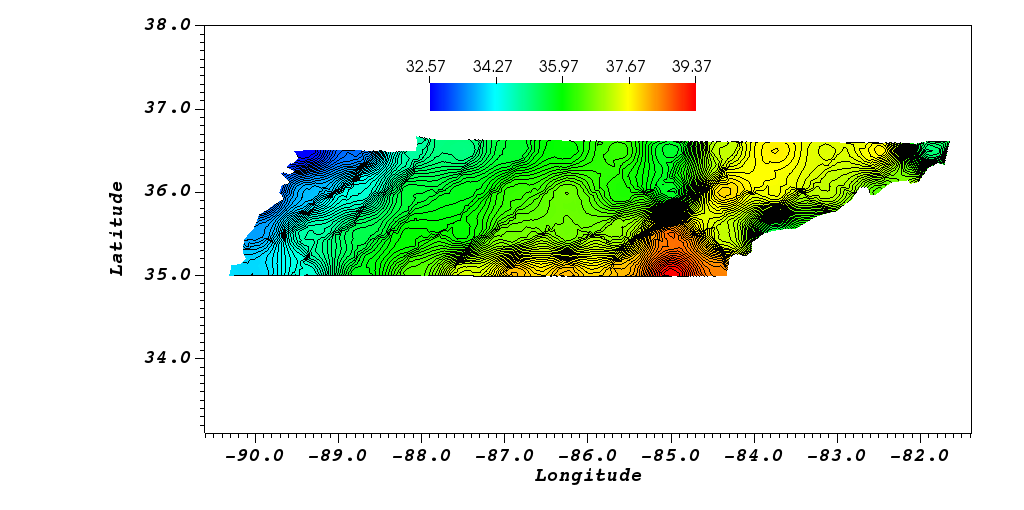}  
	\caption{Gravity Model: Hour 4}
	\label{fig:gravity4}
\end{figure}

\begin{figure}
	\centering
	\includegraphics[trim=0 0 0 0,clip, width=\textwidth]{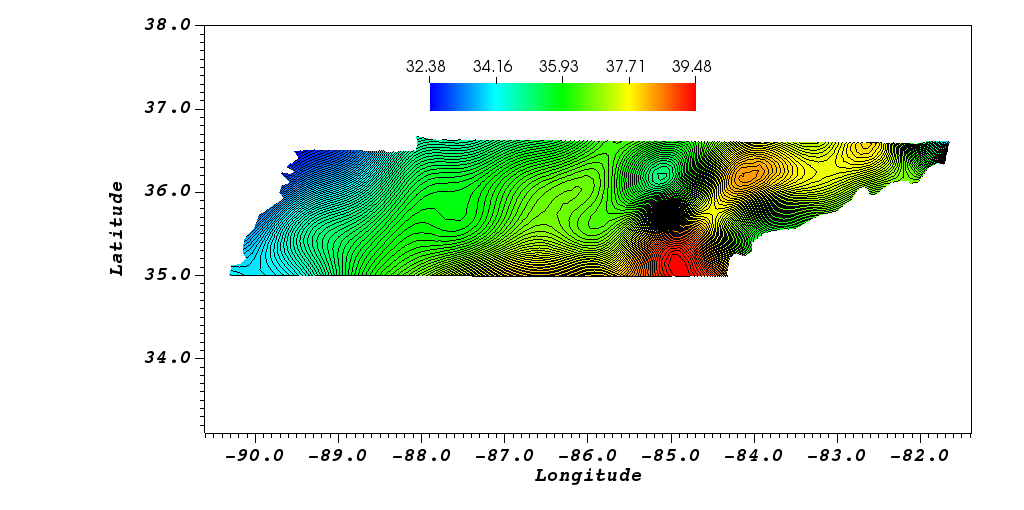}  
	\caption{RBF: Hour 4}
	\label{fig:rbf4}
\end{figure}

\begin{figure}
	\centering
	\includegraphics[trim=0 0 0 0,clip, width=\textwidth]{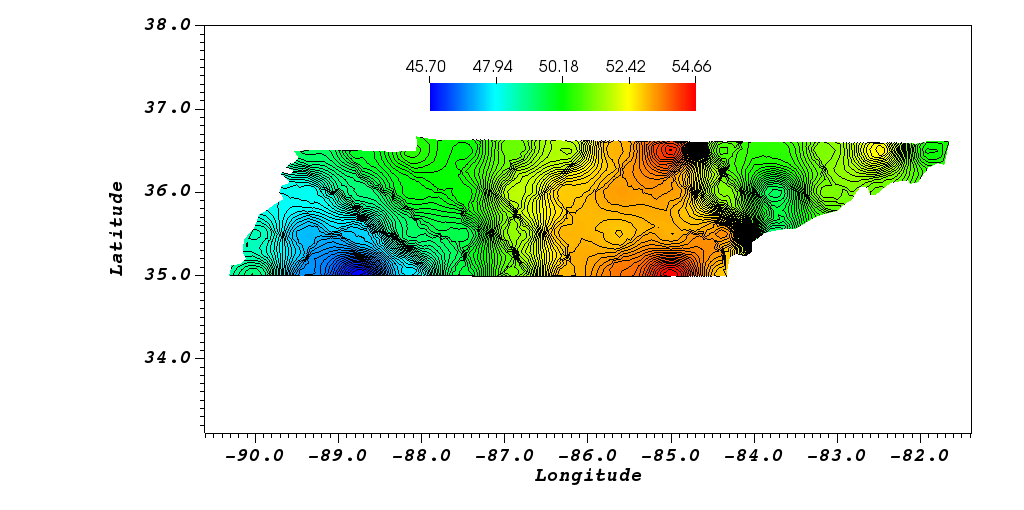}  
	\caption{Gravity Model: Hour 5}
	\label{fig:gravity5}
\end{figure}

\begin{figure}
	\centering
	\includegraphics[trim=0 0 0 0,clip, width=\textwidth]{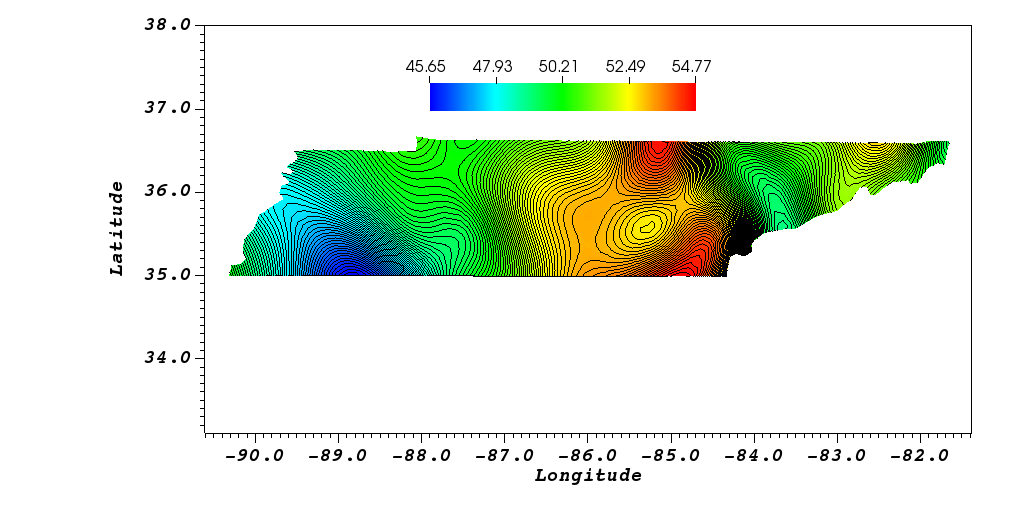}  
	\caption{RBF: Hour 5}
	\label{fig:rbf5}
\end{figure}

\begin{figure}
	\centering
	\includegraphics[trim=0 0 0 0,clip, width=\textwidth]{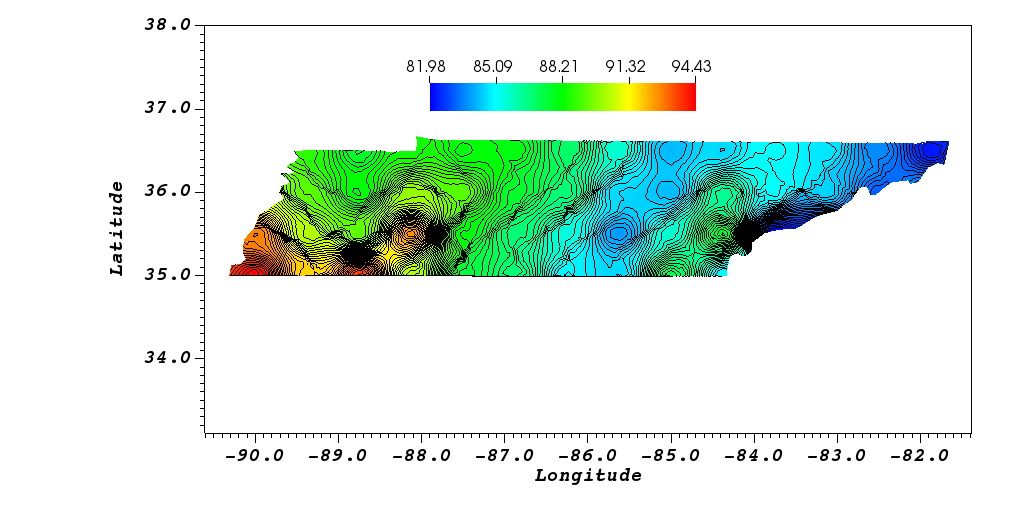}  
	\caption{Gravity Model: Hour 6}
	\label{fig:gravity6}
\end{figure}

\section{Conclusions}

Using the gravity model, it can be concluded that the number of data used for interpolation plays a role in increasing the accuracy of the output regardless of how close the data is to the point of interest. The interpolation considering all data points (Case 1) provided more accurate results that those which that considered a fixed number of points (Case 2). When comparing the interpolation outputs with the  actual analytical values, the RBF model showed a more accurate prediction than the gravity model. Apart from showing more accuracy, the RBF model achieve its accuracy faster with a smaller number of data points as compared to the gravity model (increase in the number of points increases the accuracy in a steady trend). As a demonstration of the accuracy in the application of the two models, MERRA climatic data was used in the state of Tennessee considering temperature values for interpolation. Contour plots developed based on the interpolation outputs of the two model. The RBF model contours were smoother and well defined than the gravity mode and the  model produced a wider range of values, with lower minimums and higher maximums than the gravity model.

%
\bibliography{ascexmpl-new}

\end{document}